\pdfoutput=1

\documentclass[11pt]{article}

\usepackage{acl}

\usepackage{times}
\usepackage{latexsym}

\usepackage[T1]{fontenc}

\usepackage[utf8]{inputenc}

\usepackage{microtype}

\usepackage{inconsolata}
\usepackage{float}
\usepackage{graphicx}
\usepackage{subcaption} 
\usepackage{adjustbox}
\usepackage{tabularx}
\usepackage{booktabs}
\title{Exploring Gender Disparities in Automatic Speech Recognition Technology}

\author{Hend ElGhazaly, Bahman Mirheidari, Nafise Sadat Moosavi, Heidi Christensen\\
        Computer Science, University of Sheffield, Sheffield, United Kingdom\\
  }

\begin{document}
\maketitle
\begin{abstract}
This study investigates factors influencing Automatic Speech Recognition (ASR) systems' fairness and performance across genders, beyond the conventional examination of demographics. Using the LibriSpeech dataset and the Whisper small model, we analyze how performance varies across different gender representations in training data. Our findings suggest a complex interplay between the gender ratio in training data and ASR performance. Optimal fairness occurs at specific gender distributions of  rather than a simple 50-50 split. Furthermore, our findings suggest that factors like pitch variability can significantly affect ASR accuracy. This research contributes to a deeper understanding of biases in ASR systems, highlighting the importance of carefully curated training data in mitigating gender bias.

\end{abstract}

\section{Introduction}

Speech technologies have demonstrated their effectiveness and efficiency across various domains, however, they can inadvertently perpetuate biases. Bias refers to a performance gap discriminating against certain groups or individuals. For example, we consider an Automatic Speech Recognition (ASR) system biased when its recognition accuracy is much worse for one gender than the other. Despite extensive research on mitigating bias in AI, the concept of bias and fairness in speech technologies remains relatively underexplored. In speech research, studies on bias have included more emphasis on speakers' demographics in training data such as gender, race, accents, and language variants \cite{feng2024towards, zhang2022mitigating, meyer2020artie, maison2023some, koenecke2020racial}. 
This study investigates influences on gender bias in ASR systems beyond the speakers' demographics. While prior research has mainly focused on gender imbalances\cite{garnerin2021investigating,abushariah2012arabic}, our study takes a comprehensive approach by examining not only the gender ratios in the training data but also the speech content readability, semantic similarity, and pitch distribution.
Thus in this work we aim to examine the following research questions:
\begin{itemize}
    \item RQ1: How do different gender distributions in the training set affect ASR performance and fairness?
    \item RQ2: What factors other than gender ratios influence the ASR performance?
\end{itemize}
Our results reveal a nuanced relationship between gender ratios in training data and ASR performance, challenging the notion that balanced gender representation correlates with best performance. Furthermore, our findings suggest that factors like pitch variability significantly affect ASR accuracy, underscoring the importance of a holistic approach to dataset composition as a mitigation strategy.
This research enhances the understanding of gender bias in ASR systems, emphasising the importance of accounting for diverse factors to achieve fairer speech recognition.

\section{Methodology and Experimental Setup}
\label{sec:setup}
We investigate the effects of gender distribution, speech content and speakers' pitch distributions in the training set on the ASR performance. The following subsections describe the datasets and ASR model used in the experiments.

\subsection{Train and Test sets: LibriSpeech}
\label{sec:librispeech}
The LibriSpeech corpus is widely used in ASR research \citep{panayotov2015librispeech}. The dataset includes recordings of individuals reading passages from books, which cover a range of genres and topics. It contains approximately 1000 hours of transcribed English audiobooks from LibriVox. The dataset is divided into train, dev, and test subsets. There are in total 5831 chapters from 1568 books read by 2484 speakers. The audio files in the dataset are chapters read by each speaker. All chapters in the dataset are unique but can be from the same books and multiple speakers read from the same book. The speakers' names, gender, book titles, and chapter titles are provided in the dataset. Since the only speakers' demographic provided is gender (denoted as female/male), we focused on investigating gender. The authors of LibriSpeech claim that the dataset is gender balanced at speaker level and there is no overlap between the subsets. 

The LibriSpeech corpus has two test sets; ``TestClean'' and ``TestOther''. We report the evaluation results on the LibriSpeech ``TestOther'' set as it is considered a more challenging dataset because of lower speech quality and background noises, making it more representative of real-world scenarios.

Table \ref{tab:librispeech} summarizes the extracted number of chapters, hours and unique speakers per gender in the training and test sets. 
\begin{table}[htb!]
    \begin{adjustbox}{max width=\columnwidth}
        \begin{tabular}{l c c c c c c}    
        \toprule
            & \multicolumn{3}{c}{\textbf{Training}} 
            & \multicolumn{3}{c}{\textbf{TestOther}}
            \\
            \cmidrule(rl){2-4} \cmidrule(rl){5-7}
            & Chapters & Hours & Speakers & Chapters & Hours & Speakers\\
            \midrule
            \textbf{Women} & 2671 & 465.32 & 1128 & 46 & 2.65 & 17\\ 
            \textbf{Men} & 2795 & 495.73 & 1210 & 44 & 2.69 & 16\\
            \textbf{All} & 5466 & 961.05 & 2338 & 90 & 5.34 & 33\\
            \bottomrule
        \end{tabular}
    \end{adjustbox}
    \caption{Gender distribution in the original LibriSpeech training and test sets}
    \label{tab:librispeech}
\end{table}
\subsection{Data Pre-processing}
\label{sec:subsets}
We used the LibriSpeech corpus, described in Section \ref{sec:librispeech}, to create 11 training subsets with varying gender distributions to investigate how increasing the number of women in the training set affects the ASR performance. The women's percentage in the training set was changed incrementally from zero to 100. Only the number of audio files with female speakers changed in each training set. The number of audio files corresponds to the number of chapters in Table \ref{tab:librispeech}, ensuring the content is different in each sample. 
As gender was the only speaker's metadata provided, we evaluated the ASR performance based on the reported speakers' gender (female/male) in the test set. Our focus on these two genders is only driven by their representation in the dataset used for our investigation. The same test set was used in all experiments to assess the performance with the same speakers. Table \ref{tab:expData} outlines information on the gender distributions of audio samples used in each training subset and test set. 
 \begin{table}[htb!]
    \centering
     \begin{adjustbox}{max width=\columnwidth}
        \small
        \begin{tabular}{l c c c}    
        \toprule
            \textbf{Dataset Composition}
            & \textbf{Women}
            & \textbf{Men}
            & \textbf{Total Audio Files}\\
            \midrule
            Original train set & 2671 & 2795 & 5466 \\ 
            \midrule
             0\%women & 0 & 2671 & 2671\\
             10\%women & 267 & 2404 & 2671\\
             ... & ...  & ... & ...\\
             90\%women & 2404 & 267 & 2671\\
             100\%women & 2671 & 0 & 2671\\
             \midrule
             TestOther & 44 & 46 & 90\\
            \bottomrule
        \end{tabular}
        \end{adjustbox}
    \caption{Number of audio files in training and test sets.}
    \label{tab:expData}
\end{table}
\subsection{ASR System: Whisper Model}
We used the pre-trained Whisper small model \cite{radford2022whisper} as a baseline in our experiments; given its high performance and widespread use. While larger models potentially offer higher accuracy, they require significantly more processing time and computational power, which exceed our available resources. Thus, the small model was chosen to ensure feasibility within these constraints. 
The model was first fine-tuned using the original LibriSpeech training set and dev set for validation, with a focus on hyperparameter optimization. We specifically targeted the number of optimization steps that run during training, learning rate, and train batch size. After experimentation, the optimal settings were determined based on the minimum WER on the dev set, which are 3000 steps, a 5.00E-05 learning rate and a batch size of 16. These settings were used in all the fine-tuning experiments.


\subsection{Evaluation Metrics}
We assessed the ASR model performance by measuring the Word Error Rate (WER) \cite{klakow2002testing}; the lower the WER the better the performance.
To evaluate the model's fairness, we measured the gender bias, by computing the difference in WER between women and men speakers, as described in \citet{feng2024towards} work.

Besides the gender ratio in the training set, we also explored other factors including text content and speakers' pitches to assess their potential impact on the results. For the text analysis, we measured the text difficulty in the train and test sets using the Flesch-Kincaid Grade Level readability metric. We also computed the cosine distance between the text in the training and test sets to measure the semantic similarity between them. We checked if the model trained with more similar text content to the test set would show better results. 

To gain a better understanding of the pitch diversity among the speakers in the datasets, we extracted the pitch from samples of chapters read by each speaker using ``parselmouth'' library (Praat in Python). The frequency ranges $F0_{min}$ and $F0_{max}$ were set to 50Hz and 600Hz respectively \citep{re2012preferences}. We then compared the pitch values in each training subset. We expect the model trained with more pitch diversity would be able to handle a wider range of speakers and hence perform better. 

\section{Results: Factors influencing the ASR performance}
We start by computing the WER of the ASR models trained with each training subset described in Section\ref{sec:subsets}. For each model, we examine the impact of increasing the audio files of women in the training set on the model's performance (WER) per gender. Next, we explore the analysis of the other potential influencing factors aside from gender ratio, i.e. text difficulty, semantic similarity, and pitch distribution. 
\subsection{Gender distribution in the training set}
\begin{figure}[ht]
\centering
  \includegraphics[width=\columnwidth]{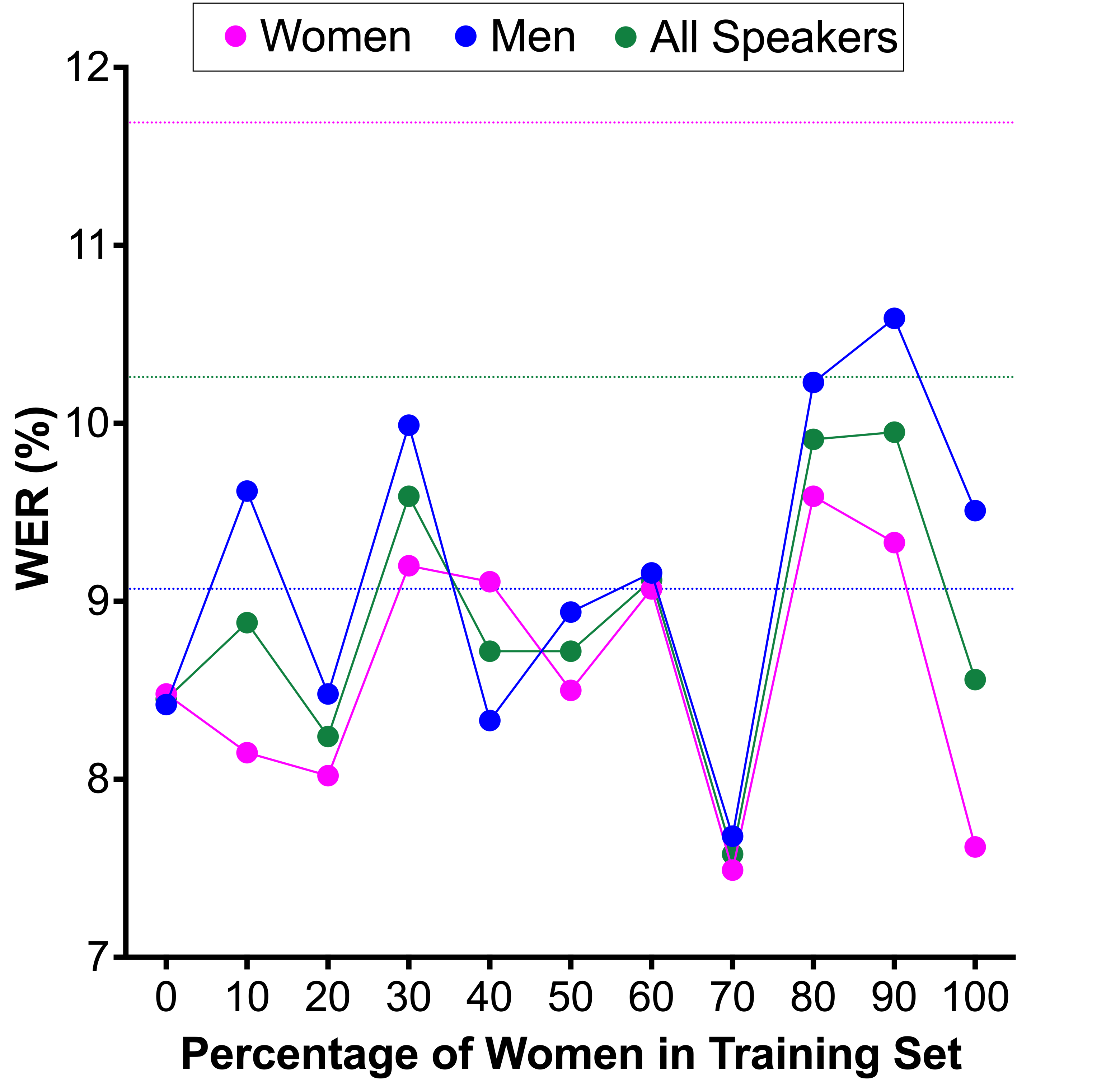}
  \caption{Mean WERs of TestOther evaluation across the fine-tuned models. The dashed horizontal lines show the results from the original Whisper model. }
  \label{fig:meanWERs}
\end{figure}

We explored how imbalances in the gender distribution in the training set affect the recognition accuracy and performance gaps between genders.
Table \ref{tab:libresults} compares the performance of each model trained with different gender distributions, starting with no women in the training set and progressively adding 10\% more women until the dataset consists entirely of women. Figure \ref{fig:meanWERs} visualizes the general trends in the results.
\begin{table}[htb!]
    \begin{adjustbox}{max width=\columnwidth}
        \begin{tabular}{l c c c c}    
        \toprule \textbf{Training Set Composition}
            & \multicolumn{3}{c}{\textbf{WER (\%)}} 
            & {\textbf{WER Gap}}
            \\
            \cmidrule(rl){2-4} 
            & Women & Men & All Speakers \\
            \midrule
            \textbf{Original Whisper model} & 11.69 & 9.07 & 10.26 & -2.62 \\
            \midrule
            \textbf{0\%women}& 8.48&8.42 &8.45& \textbf{-0.06} \\
            \textbf{10\%women}& 8.15&9.62 &8.88 &1.47 \\
             \textbf{20\%women}&8.02 &8.48 &8.24&0.46 \\
              \textbf{30\%women}&9.20 & 9.99& 9.59 &0.79 \\
               \textbf{40\%women}&9.11 &8.33 &8.72 &-0.78 \\
                \textbf{50\%women}&8.50 &8.94 &8.72 &0.44 \\
                 \textbf{60\%women}&9.07 &9.16 &9.12 &0.09 \\
                  \textbf{70\%women}&\textbf{7.49}&\textbf{7.68} &\textbf{7.58} &0.19 \\
                   \textbf{80\%women}&9.59 &10.23 &9.91 &0.64 \\
                   \textbf{90\%women}&9.33 &10.59 &9.95 &1.26 \\
                   \textbf{100\%women}&7.62 &9.51 &8.56 &1.89 \\
            \bottomrule
        \end{tabular}
    \end{adjustbox}
    \caption{Evaluation results on TestOther test set. The (-) in WER gap denotes that the women's WER is higher, i.e. worse performance with women speakers}
    \label{tab:libresults}
\end{table}

Our analysis of the ASR performance reveals that the baseline Whisper small model exhibits a gender bias, with a higher WER for women (11.69\%) compared to men (9.07\%), resulting in a notable WER gap of 2.62\%. Fine-tuning the Whisper model in general improved the performance and reduced the gender bias by up to 97\%. 
Training on a dataset with no women yields the smallest WER gap (-0.06\%), yet it is not the best performance for either gender. Increasing the proportion of women in the training set generally improves WER for female speakers, with the best overall WER (7.58\%) observed at 70\% women representation. However, extreme imbalances in gender representation, such as 10\% (WER gap: 1.47\%) or 100\% women (WER gap: 1.89\%), amplifies the WER disparities. We also found that the model trained with 90\% women has the worst WER for men (10.59\%). The men's WERs generally suggest that the model’s performance with men’s speech is negatively impacted as the women in the training set increase.
The smallest positive WER gap (0.09\%) occurs at 60\% women, suggesting that this ratio provides the best balance between accuracy and fairness.

These findings show that the original Whisper model is biased against women, but fine-tuning with a more balanced speaker distribution can help mitigate this bias. 
However, the results do not show a direct linear correlation between the gender ratio in the training set and WER. Instead, the relationship appears to be nonlinear and complex, with fluctuations in WER depending on the gender distribution. The inconsistency patterns of the WERs suggest that the gender ratio in the training set might not be the only contributing factor. Therefore, we investigated additional factors; the text complexity and pitch distributions in the datasets.
\begin{figure*}[t]
\centering   \includegraphics[width=0.48\linewidth]{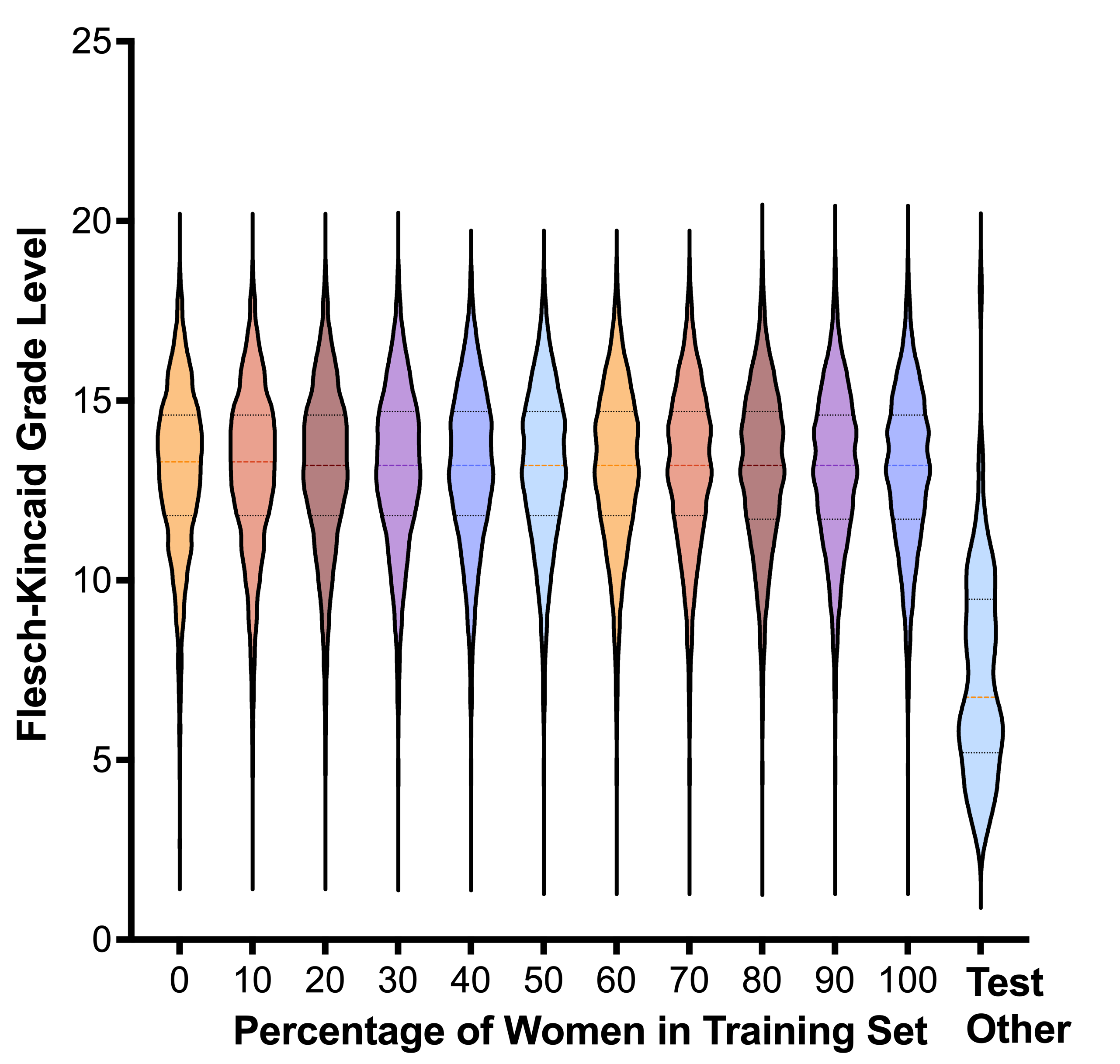} \hfill
\includegraphics[width=0.48\linewidth]{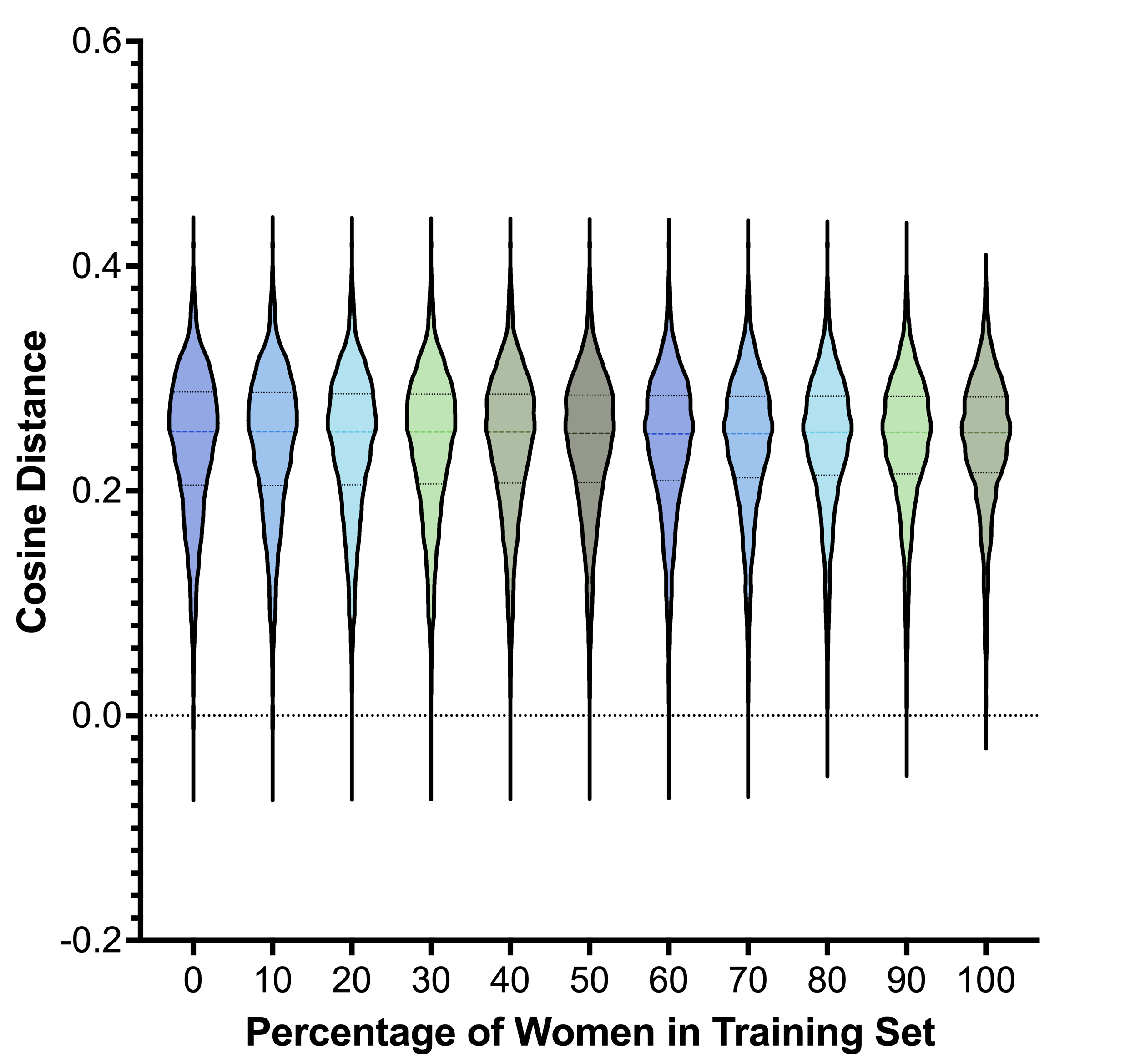}
  \caption {Comparisons between training and test sets' text difficulty (left) and semantic similarity (right).}
\label{fig:textanalysis}
\end{figure*}
\subsection{Text content in the training and test sets}
As the gender distribution in the training set did not have a direct correlation with the WER, we analyzed the text read by the speakers to investigate its impact on the ASR performance and fairness.
Text difficulty may influence the speaker's pronunciation, clarity, and fluency, all of which could impact the ASR accuracy. More complex or challenging texts might introduce variations in speech that are harder for the ASR system to process effectively. Similarly, the semantic similarity between the text read by the speaker in the test set and the training set can have an impact. If the content of the spoken text is semantically dissimilar from the training data, the ASR system may struggle to recognize certain words or phrases, leading to lower accuracy. 
Figure \ref{fig:textanalysis} represents the results of difficulty scores based on Flesch-Kincaid Grade Level readability measurement calculated for each chapter in the training and test sets. Our goal was to investigate whether varying text difficulty contributes to differences in WER across experiments.
From Figure \ref{fig:textanalysis}, it is evident that there is no significant disparity in text readability among chapters across different training subsets in our experiments. This could be because all the text is from formal English books and therefore would be at a similar difficulty level. However, the chapters in the test data appear to be easier than those in the training data. Consequently, differences in text difficulty between training subsets are not a key factor contributing to performance variations.
The second figure on the right in Figure \ref{fig:textanalysis} illustrates the semantic similarity between the content of the chapters read in each training subset and those of the TestOther. Likewise, we noticed that the shapes and ranges of the violin plots remain consistent across different training subsets. This consistency suggests that the semantic similarity of the chapters in the training data to those in the test data does not significantly impact the corresponding WER.

\subsection{Pitch distributions of the speakers}
\begin{figure*}[ht]
  \includegraphics[width=\linewidth]{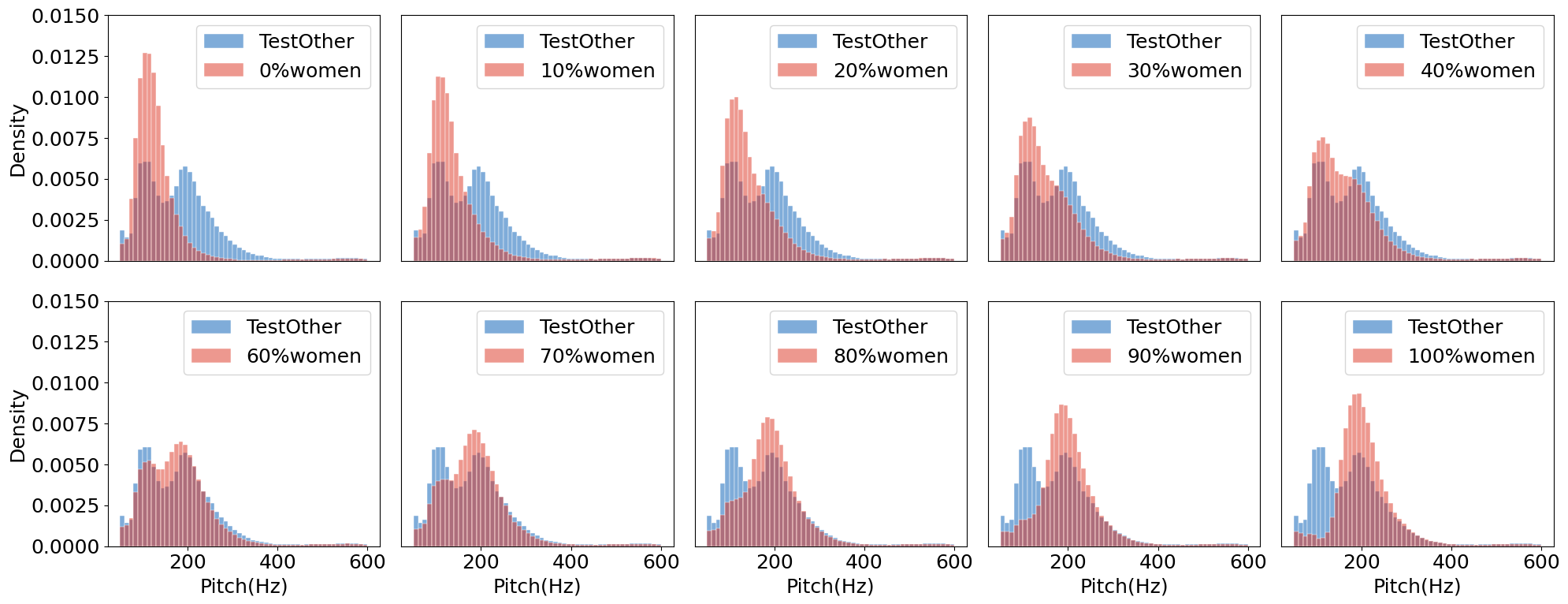}
  \caption {Illustrations of the pitch distributions in the training subsets (in red) and TestOther set (in blue).}
  \label{fig:wperPitch}
\end{figure*}
Figure \ref{fig:wperPitch} shows the pitch distributions per gender in each training set used to train the models. We hypothesised that the training sets with pitch distribution similar to the test sets would have better performance.
The plots show that more balanced training sets cover more distributions and are relatively similar to the test sets. This could explain the lower WER of the models trained with these training sets. On the other hand, there is less diversity in the higher pitch ranges in the imbalanced men-dominant training sets. This suggests that the model's higher WER (e.g. 8.88\%)  was due to its inability to handle the higher pitch ranges in the test set (Figure \ref{fig:meanWERs}). Similarly, the models trained with mostly women's voices included less frequency of lower pitch ranges and had the highest WER (e.g. 9.95\%). The findings from the pitch analysis suggest that having a greater diversity of speakers' speech attributes contributes to better WERs. 

\section{Discussion and Conclusions}
Our results showed that there is no simple direct correlation between gender ratio and WER. Instead, the results suggest that optimal performance and fairness occur at specific gender distributions (e.g.70\% women) rather than a straightforward linear trend. This highlights the importance of carefully selecting training data distributions rather than assuming that balance always leads to better results.
Training with extremely unbalanced datasets (10\% or 100\% women) can lead to less fairness, as seen in the WER gaps in Table \ref{tab:libresults}. The best trade-off between accuracy and fairness appears to be around 60-70\% women, where the WER remains low and the WER gap is minimized (Figure \ref{fig:wperPitch}). It is therefore important to consider the overall WER when mitigating gender bias and choose a model that minimises WER for different genders without compromising the overall performance.

While gender representation in training data is a significant factor, our study revealed that it should not be the sole focus. Instead, a combination of diverse factors must be considered to enhance the fairness and performance of ASR. 
This echoes the insights from researchers such as \citet{meng2022don}, who highlight the importance of content and prosody. We measured the readability of the text read by the speakers, the semantic similarity between the text in the training and test data, and pitch distributions in both training and test data. Our investigation into the text difficulty and semantic similarity suggest that these elements do not directly affect the WER in various settings (Figure \ref{fig:textanalysis}). Additionally, our analysis of pitch distributions brings to light significant differences contingent upon the pitches in the training data, emphasizing the need for a balanced representation to encompass a diverse range of pitches regardless of gender (Figure \ref{fig:wperPitch}). 

While this study focuses on an English dataset, linguistic features such as morphology, syntax, and word embeddings can vary across languages. Future work could explore whether the observed patterns hold across other languages and ASR models.

\bibliography{custom}

\end{document}